# Is Neural Machine Translation Ready for Deployment? A Case Study on 30 Translation Directions

*Marcin Junczys-Dowmunt[1,2], Tomasz Dwojak[1,2], Hieu Hoang[3]*

[1]Faculty of Mathematics and Computer Science, Adam Mickiewicz University in Poznań
[2]School of Informatics, University of Edinburgh
[3]Moses Machine Translation CIC
{junczys,t.dwojak}@amu.edu.pl hieu@hoang.co.uk

## Abstract

In this paper we provide the largest published comparison of translation quality for phrase-based SMT and neural machine translation across 30 translation directions. For ten directions we also include hierarchical phrase-based MT. Experiments are performed for the recently published United Nations Parallel Corpus v1.0 and its large six-way sentence-aligned subcorpus. In the second part of the paper we investigate aspects of translation speed, introducing AmuNMT, our efficient neural machine translation decoder. We demonstrate that current neural machine translation could already be used for in-production systems when comparing words-per-second ratios.

## 1. Introduction

We compare the performance of phrase-based SMT, hierarchical phrase-based, and neural machine translation (NMT) across fifteen language pairs and thirty translation directions. [1] recently published the United Nations Parallel Corpus v1.0. It contains a subcorpus of ca. 11M sentences fully aligned across six languages (Arabic, Chinese, English, French, Russian, and Spanish) and official development and test sets, which makes this an ideal resource for experiments across multiple language pairs. It is also a compelling use case for in-domain translation with large bilingual in-house resources. We provide BLEU scores for the entire translation matrix for all official languages from the fully aligned subcorpus.

We also introduce AmuNMT[1], our efficient neural machine translation decoder and demonstrate that the current set-up could already be used instead of Moses in terms of translation speed when a single GPU per machine is available. Multiple GPUs would surpass the speed of the proposed in-production Moses configuration by far.

[1]https://github.com/emjotde/amunmt

## 2. Training data

### 2.1. The UN corpus

The United Nations Parallel Corpus v1.0 [1] consists of human translated UN documents from the last 25 years (1990 to 2014) for the six official UN languages, Arabic, Chinese, English, French, Russian, and Spanish. Apart from the pairwise aligned documents, a fully aligned subcorpus for the six official UN languages is distributed. This subcorpus consists of sentences that are consistently aligned across all languages with the English primary documents. Statistics for the data are provided below:

| Documents | Lines | English Tokens |
|---|---|---|
| 86,307 | 11,365,709 | 334,953,817 |

Table 1: Statistics for fully aligned subcorpus

Documents released in 2015 (excluded from the main corpus) were used to create official development and test sets for machine translation tasks. Development data was randomly selected from documents that were released in the first quarter of 2015, test data was selected from the second quarter. Both sets comprise 4,000 sentences that are 1-1 alignments across all official languages. As in the case of the fully aligned subcorpus, any translation direction can be evaluated.

### 2.2. Preprocessing

Sentences longer than 100 words were discarded. We lowercased the training data as it was done in [1]; the data was tokenized with the Moses tokenizer. For Chinese segmentation we applied Jieba[2] first.

### 2.3. Subword units

To avoid the large-vocabulary problem in NMT models [2], we use byte-pair-encoding (BPE) to achieve open-vocabulary translation with a fixed vocabulary of subword symbols [3]. For all languages we set the number of subword units to 30,000. Segmentation into subword units is applied after any other preprocessing step. During evaluation, subwords are reassembled.

## 3. Phrase-based SMT baselines

[1] provided baseline BLEU scores for Moses [4] configurations that were trained on the 6-way subcorpus. Their configurations were the following:

The training corpora were split into four equally sized parts that were aligned with MGIZA++ [5], running 5 iterations of Model 1 and the HMM model on each part. A 5-gram language model was trained from the target parallel data, with 3-grams or higher order being pruned if they occured only once. Apart from the default configuration with a lexical reordering model, a 5-gram operation sequence model [6] (all n-grams pruned if they occur only once) and a 9-gram word-class language model with word-classes produced by word2vec [7] (3-grams and 4-grams are pruned if they occur only once, 5-grams and 6-grams if they occur only twice, etc.) were added, both trained with KenLM [8]. Significance pruning [9] was applied to the phrase-table and the compact phrase-table and reordering data structures [10] were used. During decoding with cube-pruning algorithm, stack size and cube-pruning pop limits of 1,000 were chosen. This configuration resembles the in-house translation systems deployed at the United Nations.

## 4. Neural translation systems

The neural machine translation system is an attentional encoder-decoder [11], which has been trained with Nematus [12]. We used mini-batches of size 40, a maximum sentence length of 100, word embeddings of size 500, and hidden layers of size 1024. We clip the gradient norm to 1.0 [13]. Models were trained with Adadelta [14], reshuffling the training corpus between epochs. The models have been trained for 1.2M iterations (one iteration corresponds to one mini-batch), saving every $30,000$ iterations. On our NVidia GTX 1080 this corresponds to roughly 4 epochs and 8 days of training time. Models with English as their source or target data were later trained for another 1.2M iterations (another 2 epochs, 8 days). For ensembling, we chose the last four model checkpoints.

## 5. Phrase-based vs. NMT – full matrix

In Figure 1 we present the results for all thirty language pairs in the United Nations parallel corpus for the officially included test set. BLEU results are case-insensitive, tokenized with the Moses tokenizer. For Chinese we applied Jieba for segmentation. BPE-subwords were concatenated.

Here we compare with NMT models that were trained for 4 epochs or 1.2M iterations. With the exception of fr-es, the neural system is always comparable or better than the phrase-based system. Especially in cases where Chinese is one of the languages in a language pair, the improvement of NMT over PB-SMT is dramatic with between 7 and 9 BLEU points. We also see large improvements for translations out of and into Arabic. No special preprocessing has been applied for Arabic. Improvements are also present in the case of the highest scoring translation directions, en-es and es-en.

## 6. Phrase-based vs. Hiero vs. NMT – language pairs with English

The impressive results for any translation direction involving Chinese motivated us to experiment with hierarchical phrase-based machine translation (Hiero) as implemented in Moses . Hiero has been confirmed to outperform phrase-based SMT for the Chinese-English language pair. We decided to expand our experiment with all language pairs that include English as these are the main translation directions the UN are working with. For these ten translation directions we created a hierarchical PB-SMT system with the same preprocessing settings as the shallow PB-SMT system.

To test the effects of prolonged training time, we also continued training of our neural systems for another four epochs or 1.2M iterations (2.4M in total)

[2] https://github.com/fxsjy/jieba

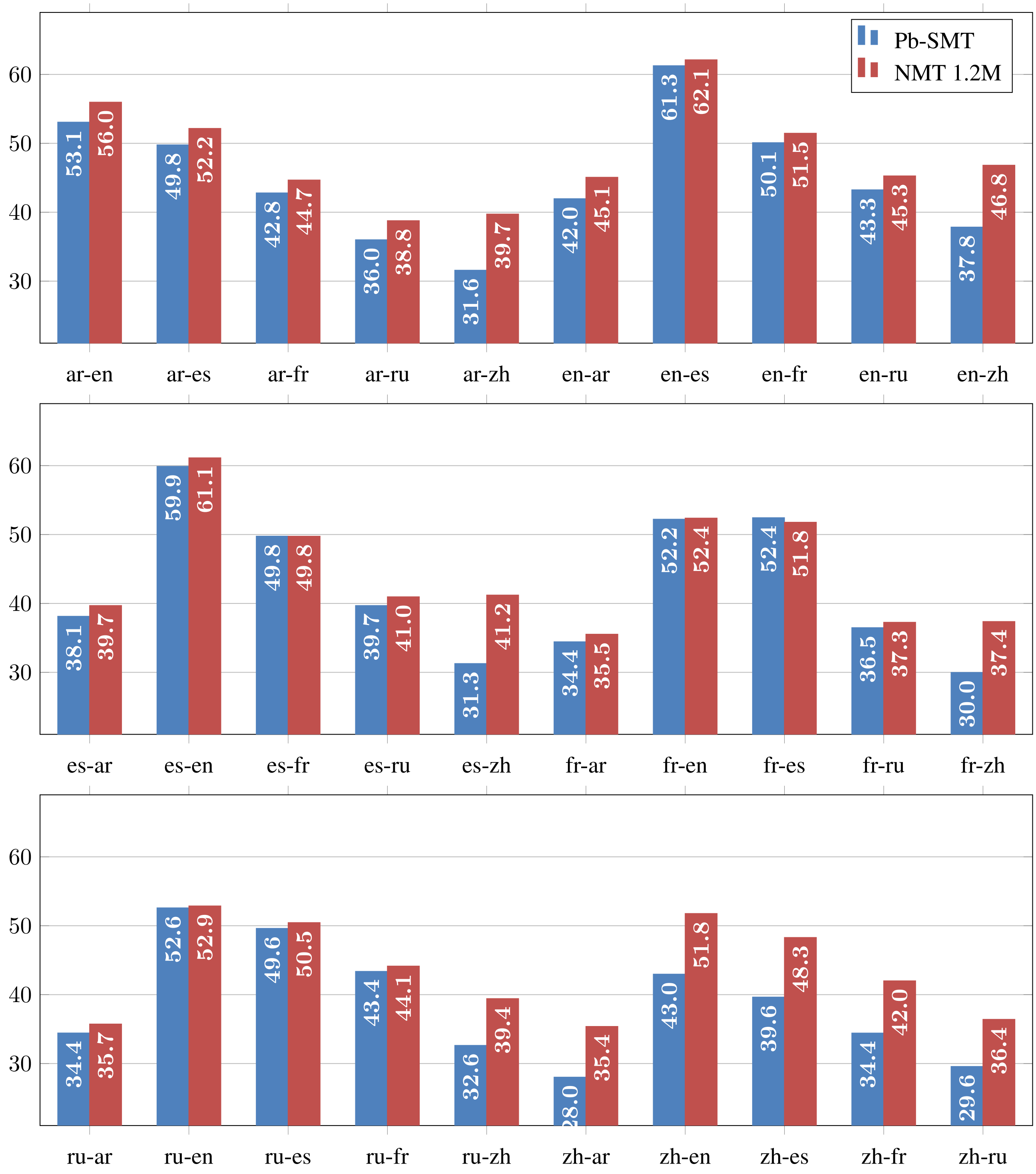

Figure 1: Comparison between Moses baseline systems and neural models for the full language pair matrix of the 6-way corpus.

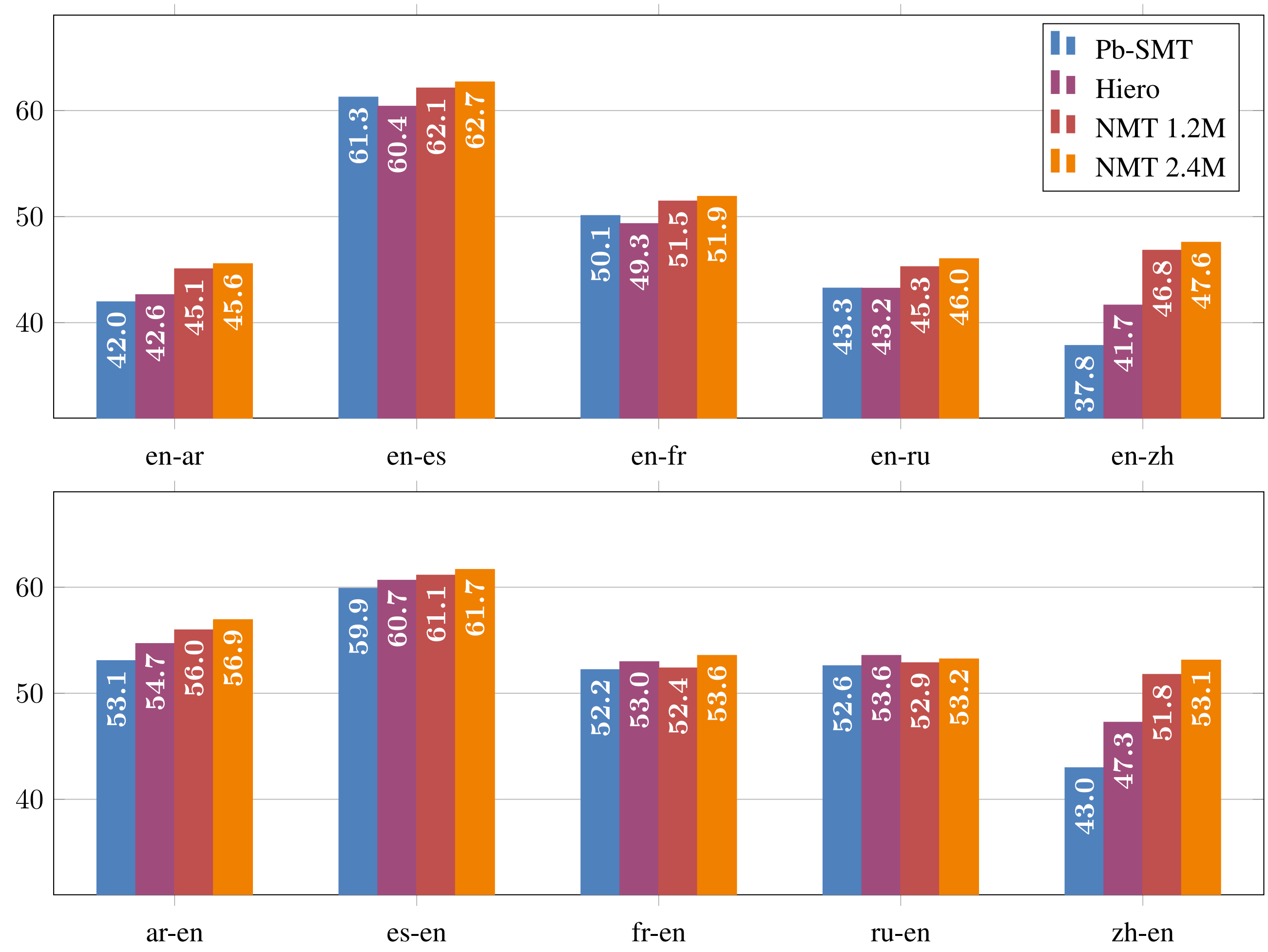

Figure 2: For all language pairs involving English, we experimented also with hierarchal machine translation and more training iterations for the neural models. NMT 1.2M means 1.2 million iterations with a batch size of 40, training time ca. 8 days. NMT 2.4 means 2.4 million iterations accordingly.

which increased training time to 16 days in total per neural system.

Figure 2 summarizes the results. As expected, Hiero outperforms PB-SMT by a significant margin for Chinese-English and English-Chinese, but does not reach half the improvement of the NMT systems. For other languages pairs we see mixed results with models trained for 1.2M iterations; for French-English and Russian-English, where results for PB-SMT and NMT are close, Hiero is the best system.

However, training the NMT system for another eight days helps, with gains between 0.3 and 1.3 BLEU. We did not observe improvements beyond 2M iterations. For our setting, it seems that stopping training after 10 days is a viable heuristic. Training with other corpus sizes, architectures, and hyper-parameters may behave differently.

## 7. Efficient decoding with AmuNMT

AmuNMT is a ground-up neural MT toolkit implementation, developed in C++. It currently consist of an efficient beam-search inference engine for models trained with Nematus. We focused mainly on efficiency and usability. Features of the AmuNMT decoder include:

- Multi-GPU, multi-CPU and mixed GPU/CPU mode with sentence-wise threads (different sentences are decoded in different threads);
- Low-latency CPU-based decoding with intra-sentence multi-threading (one sentence makes use of multiple threads during matrix operations);
- Compatibility with Nematus [12];
- Ensembling of multiple models;
- Vocabulary selection in the output layer [15, 16];
- Integrated segmentation into subword units [3].

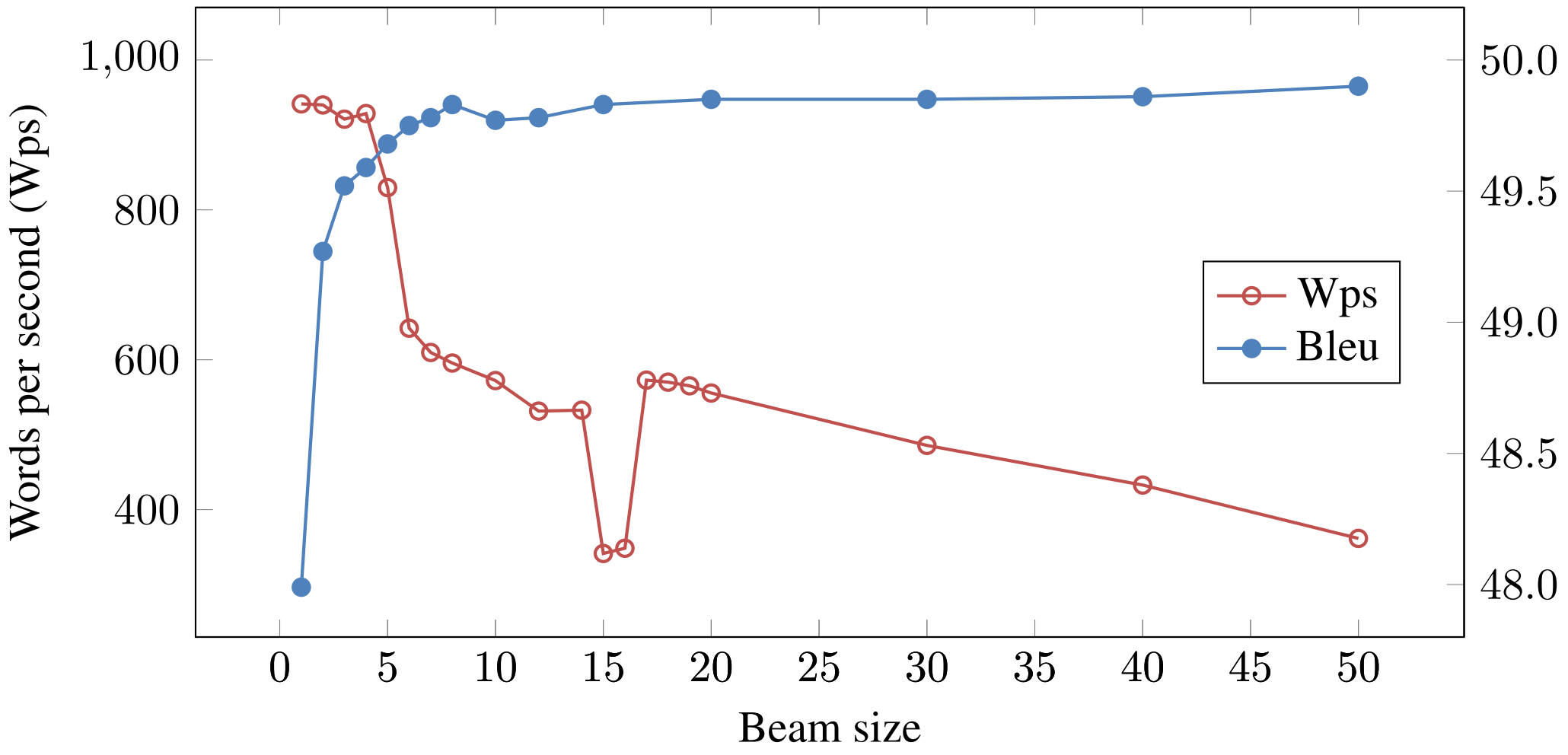

Figure 3: Beam size versus speed and quality for a single English-French model. Speed was evaluated with AmuNMT on a single GPU. The ragged appearance is due to CuBLAS GEMM kernel switching for different matrix sizes.

| System | BLEU | Wps | Memory |
|---|---|---|---|
| Single best | 49.68 | ~ 860 | 301M |
| Last-4 ensemble | 51.38 | ~ 200 | 2.1G |
| Last-4 average | 51.13 | ~ 860 | 301M |
| Last-8 average | 51.35 | ~ 860 | 301M |

Table 2: Checkpoint ensembling and averaging. Wps is translation speed as words-per-second, one GPU. Memory is memory consumption or total model size.

### 7.1. Checkpoint ensembling and averaging

Results in Figures 1 and 2 are reported for ensembles of the last four model checkpoints. Generally, this improves quality by up to 2 BLEU points, but reduces translation speed with growing ensemble size. An alternative found to work by [18] is checkpoint averaging, where a single model is produced by element-wise averaging of all corresponding model parameters, producing a single model.[3] We observed that averaging over 8 last checkpoints yields similar quality as ensembling of the last 4 checkpoints (Table 2). Results for averaging more models are uneven as increasingly weaker models are used. We did not investigate ensembles of models produced by separate training runs which would likely yield more improvements.

[3] In later experiments and discussions with Rico Sennrich, we have found that this effect seems to be an artifact of the Adadelta algorithm. Training with Adam results in better single models and seems to make checkpoint ensembling or averaging obsolete.

### 7.2. Vocabulary selection

Our vocabulary size is already greatly reduced due to the use of subword units (30k items), but especially for the CPU version of AmuNMT decoding time is dominated by the calculations in the final output layer. Previous work [15] proposed to clip the per-sentence vocabulary to the $K = 30$k (out of 500k) most common target words and $K' = 10$ most probable translations for each source word. We re-use Moses lexical translation tables trained on data segmented into subword units to obtain the translation probabilities.

Our implementation is based on [15], however, similar as [16], we find that relying mostly on $K'$ does not result in deterioration of translation quality and that $K$ can be greatly reduced. We empirically chose $K = 75$ and $K' = 75$, which results on average in ca. 1250 vocabulary items per sentence. BLEU scores remain the same although translations are slightly changed. For performance comparisons see Section 7.4.

### 7.3. Beam size vs. speed and quality

Beam size has a large impact on decoding speed. In Figure 3 we plot the influence of beam size on decoding speed (as words per second) and translation quality (in BLEU) for the English-French model. The English part of the UN test set consists or ca. 120.000 tokens, the whole test set was translated for each experiment. As can be seen, beam sizes beyond 5-7 do not result in significant improvements as the quality for a beam size of 5 is only 0.2 BLEU below the maximum. How-

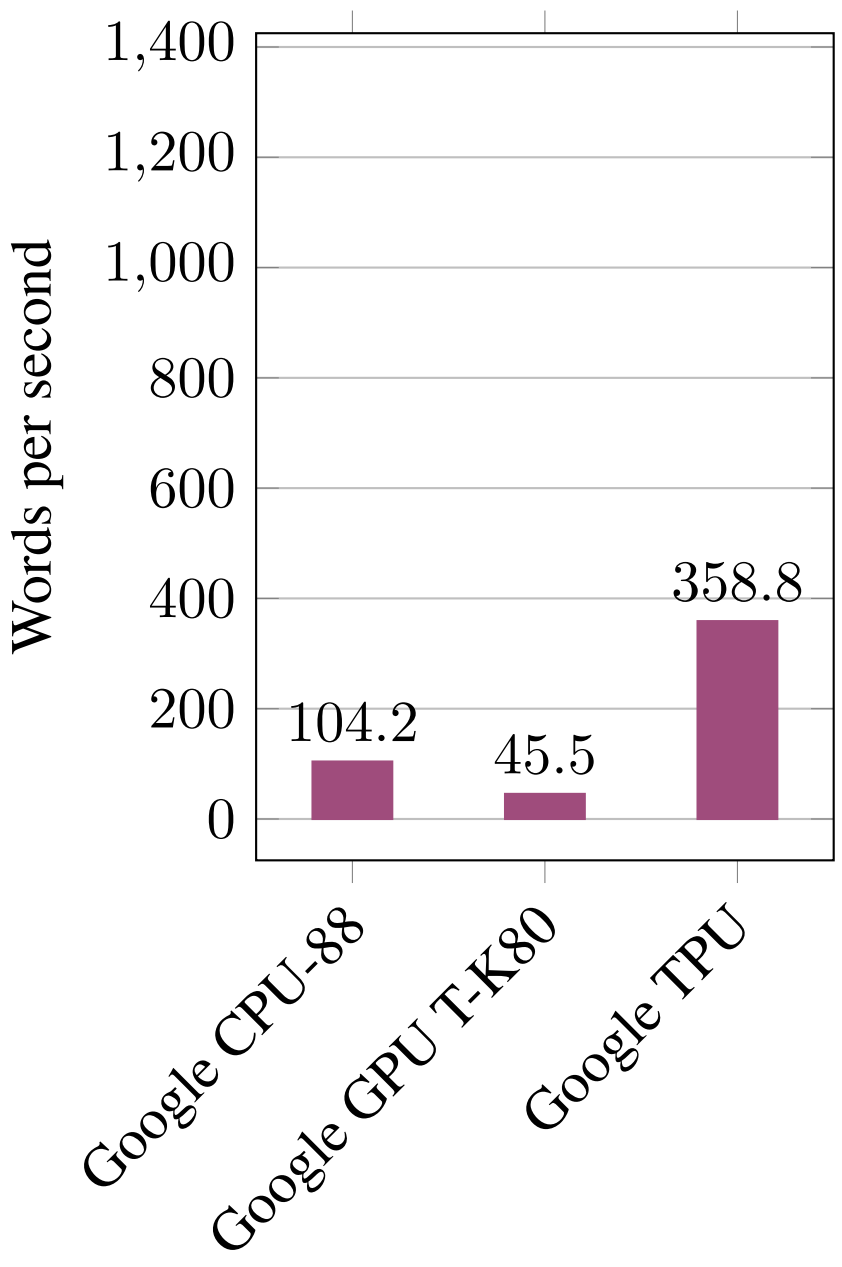

(a) Words per second for Google NMT system reported by [17]. We only give this as an example for a production-ready system.

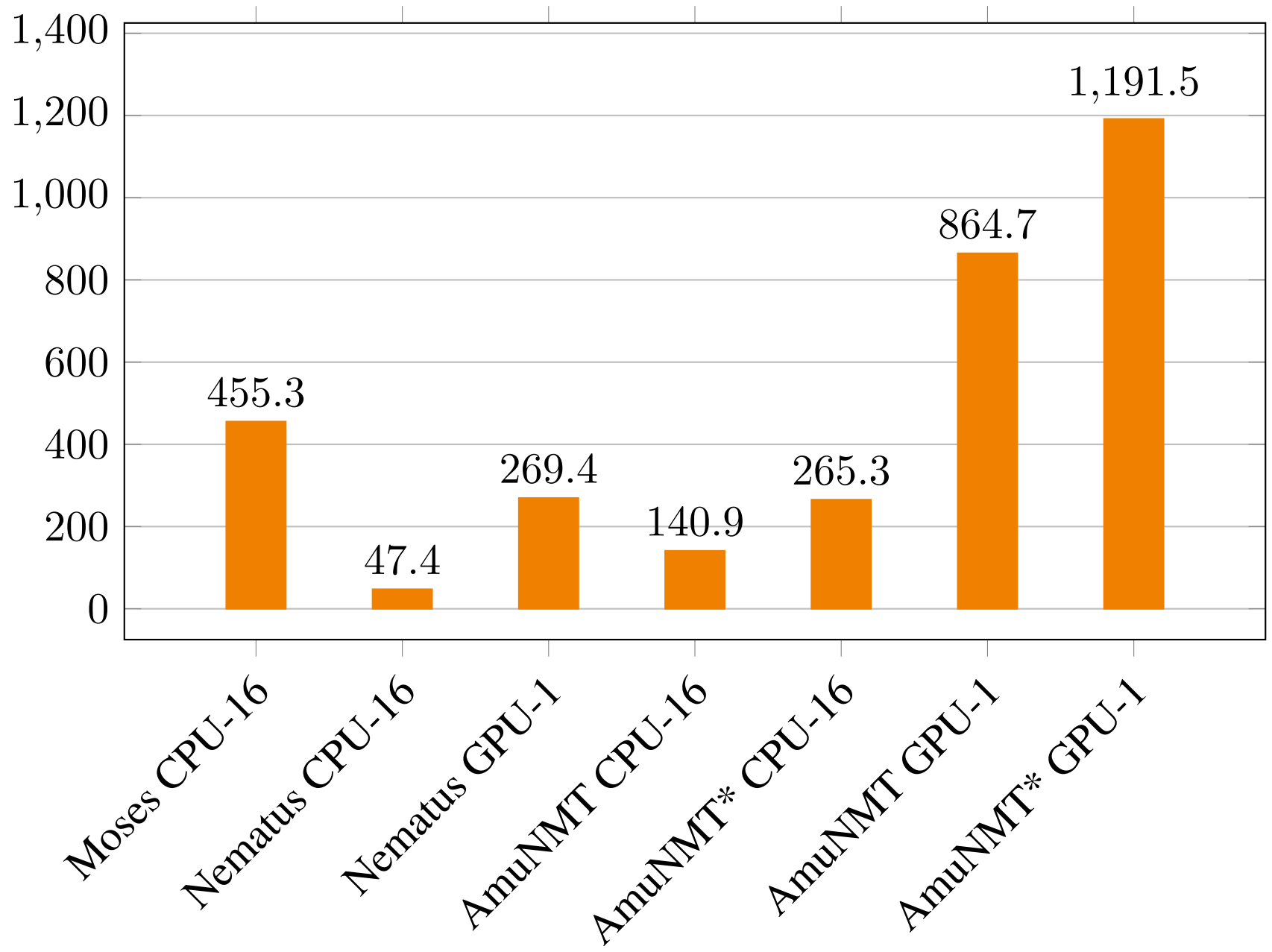

(b) Moses vs. Nematus vs. AmuNMT: Our CPUs are Intel Xeon E5-2620 2.40GHz, our GPUs are GeForce GTX 1080. CPU-16 means 16 CPU threads, GPU-1 means single GPU. All NMT systems were run with a beam size of 5. Systems marked with * use vocabulary selection.

Figure 4: Comparison of translation speed in words per second

ever, decoding speed is significantly slower. We therefore choose a beam-size of 5 for our experiments.

### 7.4. AmuNMT vs. Moses and Nematus

In Figure 4a we report speed in terms of words per second as provided by [17]. Although their models are more complex than ours, we quote these figures as a reference of deployment-ready performance for NMT.

We ran our experiments on an Intel Xeon E5-2620 2.40GHz server with four NVIDIA GeForce GTX 1080 GPUs. The phrase-based parameters are described in Section 3 which is guided by best practices to achieving reasonable speed vs. quality trade-off [19]. The neural MT models are as described in the previous section.

We present the words-per-second ratio for our NMT models using AmuNMT and Nematus, executed on the CPU and GPU, Figure 4b. For the CPU version we use 16 threads, translating one sentence per thread. We restrict the number of OpenBLAS threads to 1 per main Nematus thread. For the GPU version of Nematus we use 5 processes to maximize GPU saturation. As a baseline, the phrase-based model reaches 455 words per second using 16 threads.

The CPU-bound execution of Nematus reaches 47 words per second while the GPU-bound achieved 270 words per second. In similar settings, CPU-bound AmuNMT is three times faster than Nematus CPU, but three times slower than Moses. With vocabulary selection we can nearly double the speed of AmuNMT CPU. The GPU-executed version of AmuNMT is more than three times faster than Nematus and nearly twice as fast as Moses, achieving 865 words per second, with vocabulary selection we reach 1,192. Even the speed of the CPU version would already allow to replace a Moses-based SMT system with an AmuNMT-based NMT system in a production environment without severely affecting translation throughput.

AmuNMT can parallelize to multiple GPUs processing one sentence per GPU. Translation speed increases to 3,368 words per second with four GPUs, to 4,423 with vocabulary selection. AmuNMT has a start-up time of less than 10 seconds, while Nematus may need several minutes until the first translation can be produced. Nevertheless, the model used in AmuNMT is an exact implementation of the Nematus model.

The size of the NMT model with the chosen parameters is approximately 300 MB, which means about 24

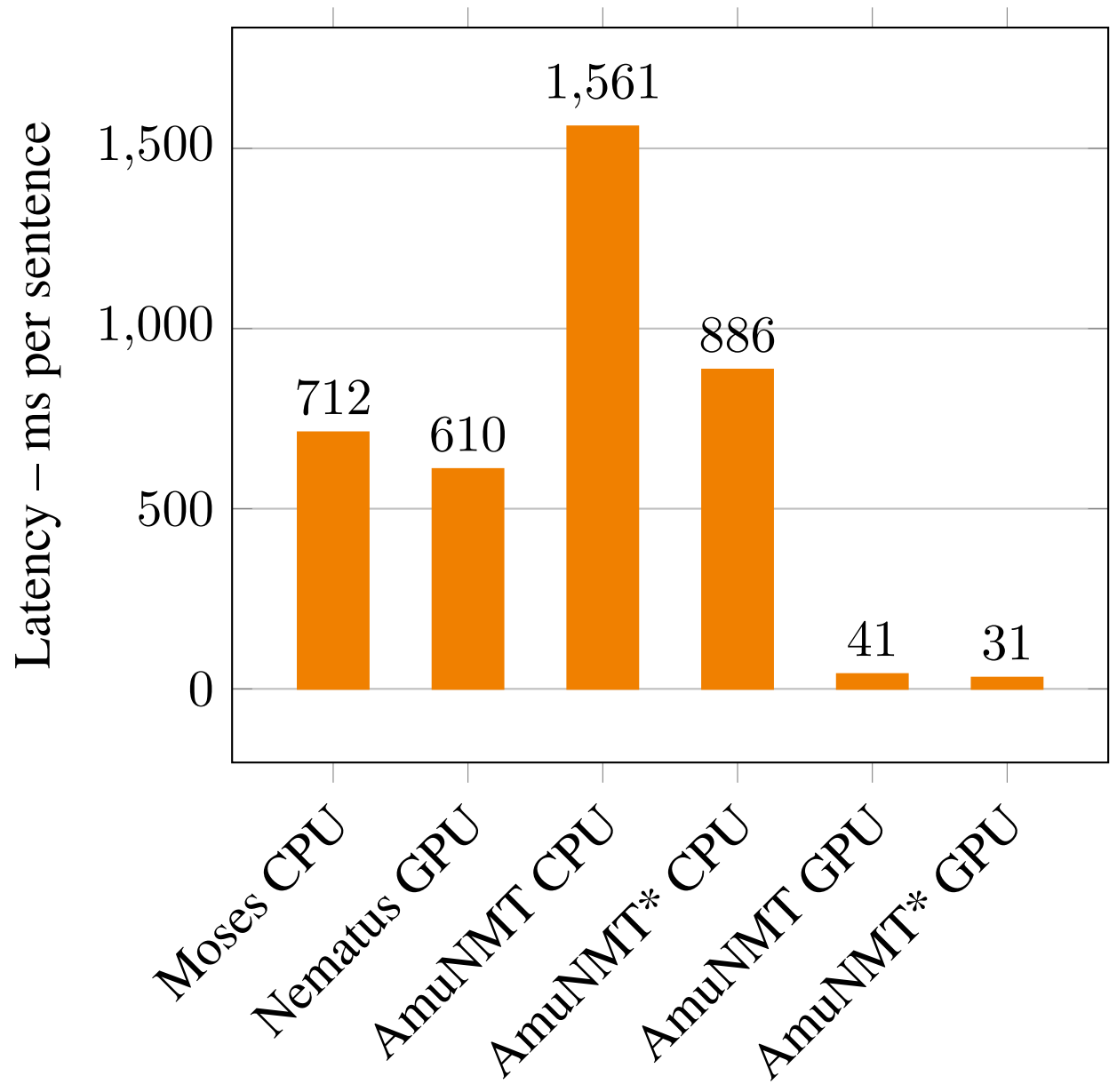

Figure 5: Latency measured in milliseconds per sentence. NMT systems were run with a beam size of 5. Lower is better. Translations were executed serially. Nematus CPU exceeded 8,500 ms and was omitted.

models could be loaded onto a single GPU with 8GB of RAM. Hardly any overhead is required during translation. With multiple GPUs, access could be parallelized and optimally scheduled in a query-based server setting.

### 7.5. Low-latency translation

Until now, we have evaluated translation speed as the time it takes to translate a large test set using a large number of cores on a powerful server. This bulk-throughput measure is useful for judging the performance of the MT system for batch translation of large number of documents.

However, there are use-cases where the latency for a single sentence may be important, for example predictive translation [20]. To compare per-sentence latency, we translate our test set with all tools serially, using at most one CPU thread or process. We do not aim at full GPU saturation as this would not improve latency. We then average time over the number of sentences and report milliseconds per sentence in Figure 5, lower values are better. Here AmuNMT GPU compares very favourably against all other solutions with a 20 times lower latency than Moses and Nematus. Latency between the CPU-only variants shows similar ratios as for bulk-translation.

## 8. Conclusions and future work

Although NMT systems are known to generalize better than phrase-based systems for out-of-domain data, it was unclear how they perform in a purely in-domain setting which is of interest for any organization with significant resources of their own data, such as the UN or other governmental bodies.

We evaluated the performance of neural machine translation on all thirty translation directions of the United Nations Parallel Corpus v1.0. We showed that for all translation directions NMT is either on par with or surpasses phrase-based SMT. For some language pairs, the gains in BLEU are substantial. These include all pairs that have Chinese as a source or target language. Very respectable gains can also be observed for Arabic. For other language pairs there is generally some improvement. In the future we would like to verify these results with human evaluation.

We introduced our efficient neural machine translation beam-search decoder, AmuNMT, and demonstrated that high-quality and high-performance neural machine translation can be achieved on commodity hardware; the GPUs we tested on are available to the general public for gaming PCs and graphics workstations. A single GPU outmatches the performance of 16 CPU threads on server-grade Intel Xeon CPUs. Access to specialized hardware seems unnecessary when planning to switch to neural machine translation with lower-parametrized models.

Even the performance of the CPU-only version of AmuNMT allows to set-up demo systems and can be a viable solution for low-throughput settings. Training, however, requires a GPU. Still, one might start with one GPU for training and reuse the CPU machines on which Moses has been running for first deployment. For future work, we plan to further improve the performance of AmuNMT, especially for CPU-only settings.

We did not cover architecture-related questions. Reducing the hidden state size by half could improve performance 4-fold. Determining the influence on translation quality would require more experiments.

## 9. Acknowledgments

This project has received funding from the European Union's Horizon 2020 research and innovation programme under grant agreement 688139 (SUMMA).

This work is sponsored by the Air Force Research Laboratory, prime contract FA8650-11-C-6160. The views and conclusions contained in this document are those of the authors and should not be interpreted as representative of the official policies, either expressed or implied, of the Air Force Research Laboratory or the U.S. Government.